\documentclass[10pt,twocolumn,letterpaper]{article}

\usepackage{iccv}              % To produce the CAMERA-READY version
\definecolor{iccvblue}{rgb}{0.21,0.49,0.74}
\usepackage[pagebackref,breaklinks,colorlinks,allcolors=iccvblue]{hyperref}
\usepackage[accsupp]{axessibility}

\usepackage{xcolor}
\usepackage{amsmath}
\usepackage{multirow}
\usepackage{booktabs}
\usepackage{amsmath}
\usepackage{amssymb}
\usepackage{float}
\usepackage[normalem]{ulem}
\usepackage{soul,color}
\usepackage{graphicx}
\usepackage{makecell} 
\usepackage{bm}    
\usepackage{newtxtext,newtxmath}

\def\paperID{8340} % *** Enter the Paper ID here
\def\confName{ICCV}
\def\confYear{2025}

\title{Translation of Text Embedding via Delta Vector \\ to Suppress Strongly Entangled Content in Text-to-Image Diffusion Models}

\author{Eunseo Koh$^{\ast}$ \;\;\; Seunghoo Hong$^{\ast}$ \;\;\; Tae-Young Kim$^{\ast}$ \;\;\; Simon S. Woo$^{\dagger}$ \;\;\; Jae-Pil Heo$^{\dagger}$ \\
    Sungkyunkwan University \\
{\tt\small \{colorrain, hoo0681, jackdawson, swoo, jaepilheo\}@g.skku.edu}
}

\newcommand\blfootnote[1]{%
  \begingroup
  \renewcommand\thefootnote{}\footnote{#1}%
  \addtocounter{footnote}{-1}%
  \endgroup
}

\begin{document}
\maketitle
\begin{abstract}
Text-to-Image (T2I) diffusion models have made significant progress in generating diverse high-quality images from textual prompts. However, these models still face challenges in suppressing content that is strongly entangled with specific words. For example, when generating an image of ``Charlie Chaplin", a ``mustache" consistently appears even if explicitly instructed not to include it, as the concept of ``mustache" is strongly entangled with ``Charlie Chaplin". To address this issue, we propose a novel approach to directly suppress such entangled content within the text embedding space of diffusion models. Our method introduces a delta vector that modifies the text embedding to weaken the influence of undesired content in the generated image, and we further demonstrate that this delta vector can be easily obtained through a zero-shot approach. Furthermore, we propose a Selective Suppression with Delta Vector (SSDV) method to adapt delta vector into the cross-attention mechanism, enabling more effective suppression of unwanted content in regions where it would otherwise be generated. Additionally, we enabled more precise suppression in personalized T2I models by optimizing delta vector, which previous baselines were unable to achieve. Extensive experimental results demonstrate that our approach significantly outperforms existing methods, both in terms of quantitative and qualitative metrics.
Codes are available at \url{https:/github.com/eunso999/SSDV}
\vspace{-10pt}

\blfootnote{
$^\ast$ Co-authors with equal contributions; listed alphabetically} 
\blfootnote{
$^\dagger$ Co-corresponding authors
}
\end{abstract}    
\section{Introduction}
% The recent advancements in the diffusion model has shown remarkable capabilities in various generative applications such as Text-to-Image (T2I) synthesis~\cite{hierarchical,photorealistic,high}. 

Recent advancements in diffusion models~\cite{denoising} have demonstrated remarkable capabilities across various generative tasks, such as Text-to-Image (T2I) synthesis~\cite{hierarchical,photorealistic,high}. However, diffusion models still face limitations in handling complex prompts. For instance, they struggle with generating a specified number of objects~\cite{countten, kang2023counting} or objects with particular attributes~\cite{liu2022compositional} as described in the prompt. These limitations highlight the limited expressiveness of diffusion models, which become especially apparent when trying to remove elements. % rather than add new ones to the output. 
For example, when given a prompt such as ``a night sky without stars", the model frequently focuses on the word ``stars", failing to exclude them from the generated image.
We refer to these undesired elements (e.g. ``star") as \textit{negative content}. 
% Due to this inherent characteristic of diffusion models, suppressing negative content, such as content potentially infringing copyright or resulting in malicious outputs, remains challenging.
Due to this inherent characteristic of diffusion models, effectively suppressing negative content remains a challenging task, making it difficult to exclude undesirable content such as copyright-infringing or malicious outputs.
\begin{figure}[t!]
    \centering
    \includegraphics[width=0.48\textwidth]{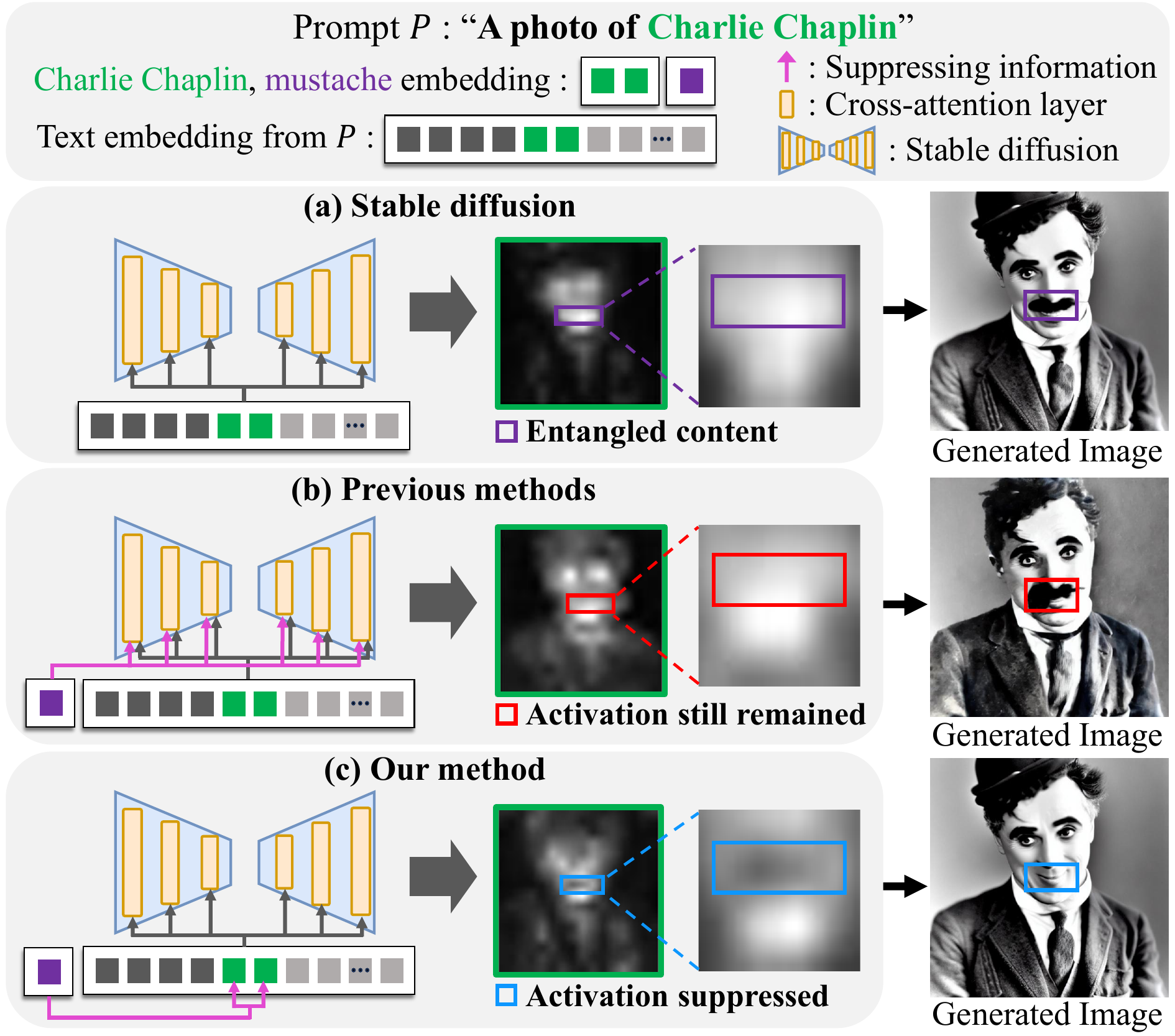} 
    \vspace{-15pt}
    \caption{(a) When generating ``A photo of Charlie Chaplin" with Stable diffusion, the cross-attention map for the word ``Charlie Chaplin" (\textbf{green border}) shows strong activation in the mustache region entangled with Charlie Chaplin. (b) Previous methods~\cite{p2p, sega, suppresseot} do not directly suppress the mustache from the embedding of ``Charlie Chaplin", which is strongly entangled with the mustache and leads to the mustache still being present.
    (c) Our method successfully suppresses the mustache by directly modifying the embedding of ``Charlie Chaplin".
    }
    \label{fig:Introduction}  
    \vspace{-13pt}
\end{figure} 

To address the aforementioned problems, several works have focused on developing methods to suppress the generation of negative content.
During the image generation process, suppression methods either control the direction of content generation~\cite{p2p, sega}, update the weights of diffusion model~\cite{erasing, forget, ablating}, or use an in-painting image to train a model to erase content via textual instructions~\cite{Instinpaint}. Also, recent work~\cite{suppresseot} proposes soft-weighted regularization to redundant negative content information from [EOT] token embedding and text embedding optimization.

Such suppression methods can effectively reduce undesired content during image generation; however, they struggle when negative content is strongly entangled with specific keywords in the input prompt.
% For example, given the input prompt ``a photo of Charlie Chaplin" with the negative content ``mustache," the model almost always generates a mustache due to its strong entanglement with ``Charlie Chaplin."
For example, given the input prompt ``A photo of Charlie Chaplin" while attempting to suppress negative content ``mustache", the model almost always generates negative content (``mustache") due to its strong entanglement with keywords (``Charlie Chaplin").
As illustrated in Fig.~\ref{fig:Introduction}.(a), the ``Charlie Chaplin" token strongly attends to the mustache region, indicating a high tendency to generate a mustache, thus making suppression challenging.
Consequently, previous methods that weaken attention to negative content tokens~\cite{p2p, suppresseot} or reduce their influence at the noise level~\cite{sega} fail to adequately suppress mustache generation associated with the ``Charlie Chaplin" token, as shown in Fig.~\ref{fig:Introduction}.(b).

To address this issue, we take inspiration from the insight that adding or subtracting a specific vector from a word embedding can strengthen or weaken certain attributes in generated images. 
We call this adjustable vector the \textit{delta vector}. Specifically, delta vector, defined in the text embedding space, is projected into the image feature space through the diffusion model’s projection layers, allowing direct control over the image.
% Therefore, when delta vector corresponds to the embedding of negative content, it affects whether negative content appears in the image. 
We define a delta vector as the embedding of negative content, enabling it to modulate whether the negative content appears in the image. 
By subtracting this vector from the original word embedding, we can effectively suppress negative content, as shown in Fig.~\ref{fig:Introduction}.(c). 
Unlike previous methods, where content strongly entangled with keywords in the input prompt hinder suppression, our method explicitly reduces the influence of these entangled content, resulting in more effective suppression.
% Unlike previous methods that faithfully render entangled concept in the input prompt, our approach explicitly reduces the influence of entangled content, resulting in more effective suppression. 
To the best of our knowledge, our work is the first to address suppressing strongly entangled content in Text-to-Image generation.

% To address this problem, we propose a method to suppress strongly entangled negative content by directly modifying the word embedding.
% By adding or subtracting a specific vector, the word embedding can be adjusted to emphasize or weaken the specific attributes.
% In this paper, we refer to this displacement vector as the \textit{delta vector}.
% The delta vector in the text embedding space is projected onto the features used for image generation, allowing it to modulate conditions in the generated image. 
% By treating the delta vector as the embedding of negative content and subtracting it from the word embedding, we effectively weaken the features responsible for generating the negative content.
% Notably, this approach enables the delta to be easily obtained in a zero-shot manner.

Furthermore, we propose Selective Suppression with Delta Vector (SSDV), a method that adaptively utilizes the delta vector within cross-attention to selectively suppress negative content while minimizing unintended effects on other image regions. 
First, we identify regions associated with negative content by adding delta vector to the word embeddings of key features, thus intensifying attention towards these regions. 
Subsequently, we subtract the delta vector from the word embeddings of value features, effectively reducing negative content within these targeted regions. 
Through quantitative and qualitative experiments, we demonstrate that our approach outperforms the baseline methods, achieving more effective suppression of strongly entangled content.

% Furthermore, 
Moreover, we show that our method effectively suppresses strongly entangled content on personalized models~\cite{dreambooth, customdiffusion}, which existing methods fail to achieve.
Additionally, to further enhance the performance on personalized models, we propose an optimization-based approach to obtain a more precise delta vector.

In summary, our work has the following contributions: 
% \SH{
\begin{itemize}
    \setlength\itemsep{1pt}
    
    \item We introduce a novel suppression method, called \textbf{Selective Suppression with Delta Vector (SSDV)}, to effectively suppress strongly entangled negative content in cross-attention operation by using delta vector.

    \item We extend our approach to personalized T2I models by introducing an optimization-based method, enabling more precise suppression of strongly entangled content, which existing methods could not achieve.

    \item Through extensive evaluations, we demonstrate that our method achieves state-of-the-art performance in both quantitative and qualitative evaluations of strongly entangled negative content suppression.
\end{itemize}
% }

% \begin{itemize}
%     \setlength\itemsep{1pt}
   
%     \item We propose a novel method to effectively suppress strongly-entangled content by introducing a delta vector that directly suppresses content from the text embedding.

%     \item We develop a method to apply the delta vector within the cross-attention mechanism for enhanced suppression quality

%     \item We introduce an optimization-based approach for the delta vector, enabling more precise suppression of strongly entangled content in personalized T2I models, which existing baselines could not achieve.
    
%     \item Our \TY{SSDV} method achieve State-of-the-Art performance 
%     % significantly outperforms existing approaches 
%     in both quantitative and qualitative evaluations of entangled content suppression.
    
% \end{itemize}
\section{Related Work}
\subsection{Text-to-Image Diffusion models}
The diffusion-based model~\cite{ generative, denoising} is a class of deep generative models based on the stochastic diffusion process. In diffusion models, a sample from the data distribution is gradually noised by the forward process, and the model learns the reverse process of gradually denoising the sample. Recently, a diffusion-based generation model guided by text prompt has shown remarkable results in Text-to-Image (T2I) generation~\cite{high, hierarchical,imagereward}. Our work is based on these T2I models, leveraging the cross-attention mechanism to suppress the generation of negative content.
% during the image generation process.
\subsection{Diffusion-Based Content Suppression}
Due to the remarkable generation performance of the T2I diffusion models, several diffusion-based methods that generate or edit images with text conditions have recently been proposed ~\cite{p2p, dragdiff, erasing, Instinpaint, suppresseot}. P2P~\cite{p2p} proposes direct control attributes in the synthesized image by various manipulations of the prompt. DragDiff~\cite{dragdiff} utilizes the feature correspondence for fine-grained image editing. With the recent development of such editing methods, content erasing is becoming increasingly important and popular in the process of controlling images. ESD~\cite{erasing} is a fine-tuning-based method that utilizes negative guidance, Inst-Inpaint~\cite{Instinpaint} trains the diffusion model using both the source image and the inpainted target image with instructional text prompts. SuppressEOT~\cite{suppresseot} proposes soft-weight regularization that suppresses negative content information on both negative content tokens and [EOT] tokens. Nonetheless, these approaches still have limitations in suppressing strongly entangled concept. In this paper, we address this issue by effectively modifying the text embedding to suppress entangled content. 
\section{Preliminary}
\subsection{Cross-Attention in Latent Diffusion Models}
Latent Diffusion Models (LDM)~\cite{high} use a U-Net architecture with iterative self-attention~\cite{attention} and cross-attention layers. Through cross-attention, LDM can incorporate various conditions, such as text and layout, enabling the synthesis of images that match the given conditions.
For our work, we use Stable Diffusion (SD)~\cite{high} conditioned on a text prompt $P$. In SD, a given text prompt $ P $ is split into multiple tokens through a tokenizer $ \tau $. Here, a single word may correspond to one or more tokens, and each token is mapped to a token embedding via the text encoder $\xi$~\cite{clip}. The text embedding $ \hat{e} \in \mathbb{R}^{M \times d}$ for a prompt $ P $ is then represented as follows:
\begin{equation}
    \label{text_embedding}
    \begin{aligned}
    % &=\hat{e}, \\
    \hat{e}=\left[e_0, e_1, e_2, \cdots, e_{M} \right]=\xi(\tau(P)),
    \end{aligned}  
\end{equation}
where $M$ denotes the number of tokens in $P$ and $ e_{0,1,2,\dots M}$ represent the embeddings corresponding to each token, with $d$ denoting the dimension of the text embedding. Consequently, the cross-attention mechanism of SD is formulated with $ \hat{e} $ as follows:
\begin{equation}
    \label{cross_attention}
    \begin{aligned}
    Q=f_Q(\phi(z_t)), \quad K = f_K(\hat{e}), \quad V = f_V(\hat{e}), \\ 
    \text{Attention}(Q, K, V) = \text{softmax}\left(\frac{Q \cdot K^\top}{\sqrt{d_k}}\right) \cdot V ,
    \end{aligned}  
\end{equation}
where ${Q}$, ${K}$, ${V}$, $d_k$, and $f(\cdot)$ denote query, key, value, dimension of the key, and projection layer, respectively. 
Similarly, $z_t$ and $\phi$ denote the latent variable and intermediate representation within the UNet of Stable Diffusion, respectively. 
As discussed in previous studies~\cite{li2023stylediffusion,p2p}, $K$ identifies regions within $Q$ where the textual condition is expected to appear, while $V$ provides the visual features representing this condition in the generated image.

% Our method leverages the token-wise competition mechanism in the cross-attention, as discussed in previous studies~\cite{li2023stylediffusion,p2p}.
% Cross-Attention between the query (image) and key (text embedding) vectors is computed along the spatial dimension, producing an \(hw \times M \) attention map, where $hw$ represents the spatial dimensions of the image or feature map.
% This attention map shows the relationship between the latent $z_t$ and the input words, with higher attention scores indicating stronger relevance. 
% The attention map is then multiplied by $V$, allowing these high-attention regions to capture more information about the input word, effectively synthesizing the word's content onto the image at the corresponding spatial locations.
%The attention map is then multiplied by $V$, enabling these highly attended locations to capture more information about the input word, thereby synthesizing the word's content onto the image at the corresponding respective locations.

\subsection{Personalized T2I model}
% To allow customized and more fine-grained image generation with pre-trained T2I model, many personalized T2I methods have been developed.
Personalized methods~\cite{dreambooth, customdiffusion, multi, jedi, hyperdreambooth, arbooth} fine-tune pre-trained T2I models to generate customized images of specific subjects or styles.
% Personalized T2I methods have been developed to customize and more fine-grained image generation with a pre-trained T2I model. 
DreamBooth~\cite{dreambooth} fine-tunes SD with a small set of subject images, enabling the generation of subject-specific images using a unique identifier token. CustomDiffusion~\cite{customdiffusion} fine-tunes the key and value weights of the cross-attention layer with a unique identifier token to enhance personalization capabilities. 
However, when fine-tuned for specific subjects or styles, both methods tend to forget the prior knowledge encoded in the original T2I model. Although they incorporate reconstruction-based loss functions to preserve this information, they still struggle to retain the original priors effectively.
% However, both methods suffer from the problem of forgetting prior knowledge of the original T2I model during fine-tuning for specific subjects or styles. Despite incorporating reconstruction-based loss functions to preserve prior information, those methods still struggle to retain the original priors effectively.
% These two methods reduce the original T2I model's prior knowledge by updating all or part of the model's parameters to align with the identifier token. 
% Although both DreamBooth and Customdiffusion include a loss function to reconstruct images for preserving prior information, they still face difficulties in maintaining the original priors. 
This weakening of prior knowledge also distorts the information associated with the content to be suppressed, making it difficult to accurately identify, thereby making the suppression process more challenging.
% Additionally, the model tends to overfit the specific subject, limiting its ability to change the attributes of this subject. This makes it challenging to suppress content entangled with the personalized subject. 
\section{Methodology}
% 중간에 기존 연구와의 차별점 강조하기
\begin{figure*}[t]
    \centering
    \includegraphics[width=0.95\textwidth]{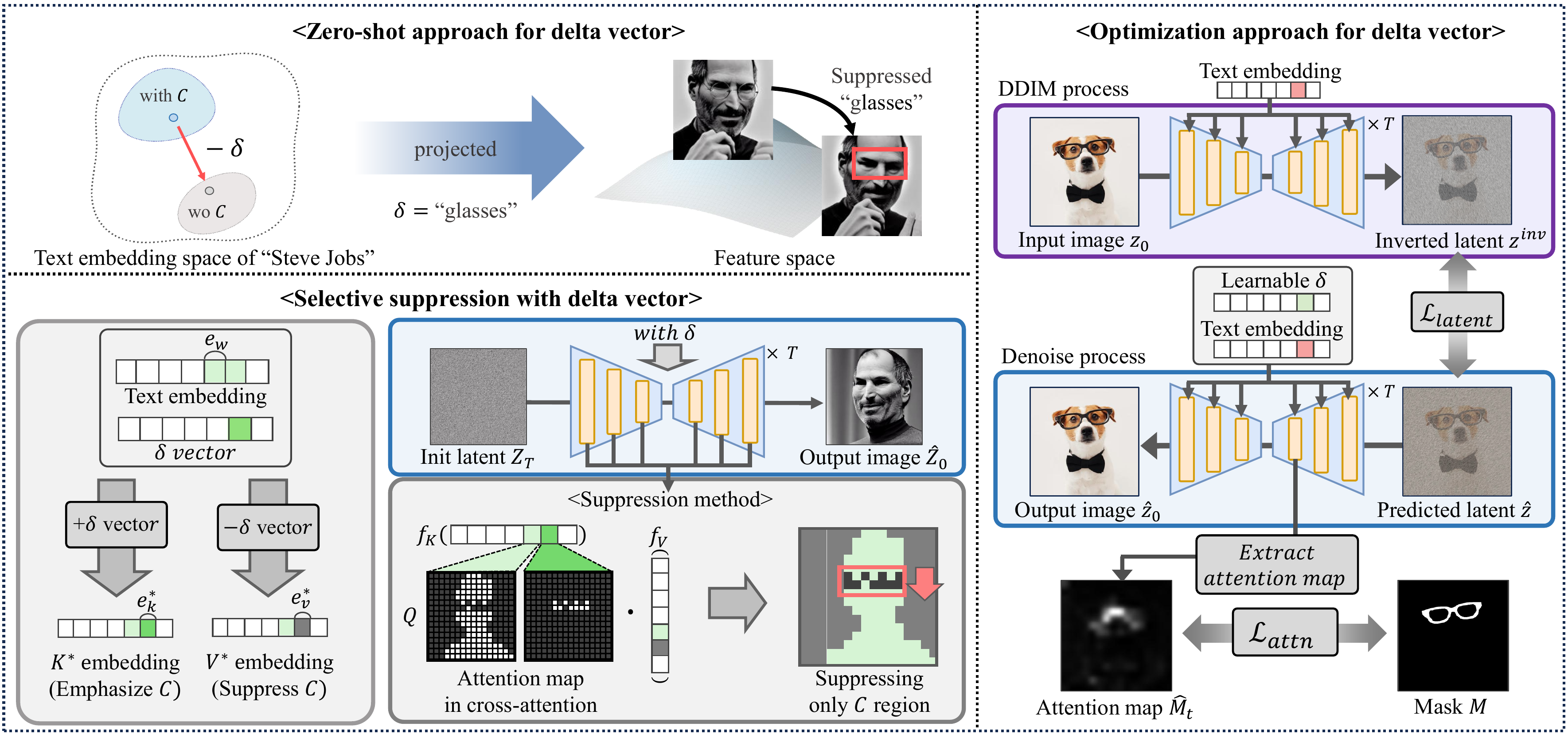} 
    \vspace{-8pt}
    \caption{Our proposed method overview. We introduce two methods for obtaining delta: a zero-shot approach that obtains delta without any additional training (top-left corner) and an optimization approach that yields a more precise delta (right side). In the bottom-left corner, we illustrate the Selective Suppression with Delta Vector, showing how the obtained delta vector is applied to the text embedding for content suppression and how it operates within the cross-attention layer.}
    \label{fig2}  
    \vspace{-10pt}
\end{figure*}

\subsection{Overview}
Our primary goal is to suppress negative content that is strongly entangled with specific keywords in the input prompt during text-to-image generation. 
To this end, we introduce the delta vector, a displacement vector in the word embedding space designed to facilitate effective suppression.
Next, we propose Selective Suppression with Delta Vector (SSDV), where the delta vector is applied to the replication tokens to create competing embeddings, enabling precise spatial suppression through softmax normalization.
Furthermore, for personalized models~\cite{dreambooth, customdiffusion}, we explore optimizing the delta vector for more precise control.

\subsection{Applying Delta Vector to Text Embedding}
In the text embedding space of T2I models such as Stable Diffusion, the embedding for the same word can vary depending on the context. 
For instance, the embedding of the word ``apple" differs depending on its context, such as between ``a photo of a green apple" and ``a photo of a red apple." 
We can use the difference between these embeddings to alter the characteristics of a specific word. 
If there is a vector that can suppress certain content, adding this vector to the embedding of a specific word can effectively reduce the presence of that content.
Motivated by this observation, we propose to suppress entangled content by introducing an embedding displacement vector, delta $\delta$.

Let $P$ be a text prompt (e.g. in Figure~\ref{fig2}, ``a photo of Steve Jobs"), where we aim for the diffusion model to generate an output that excludes a negative content $C$ (e.g. ``glasses"). We denote the word of interest in $ P $ as $ w $ (e.g. ``Steve Jobs"), and its corresponding word embedding as $e_w$. Our approach modifies the embedding $e_w$ using the delta vector $\delta$ as follows:
\begin{equation} 
\begin{aligned} 
e_{w}^{*} = e_{w} + \alpha \delta,
\end{aligned} \label{Applying-Delta}
\end{equation}
where $e_w^*$ is the translated word embedding with a modification coefficient $\alpha\in \mathbb{R}$. We construct the delta vector $\delta\in\mathbb{R}^{n_w\times768}$ by replicating a single-length vector $\delta_0\in \mathbb{R}^{1\times768}$ to match the length $n_w$ of the word $w$, thus applied to the word embedding $e_w \in \mathbb{R}^{n_w\times768}$.
% The delta vector $\delta$ lies in $\mathbb{R}^{n_w\times768}$, where $n_w$ is the length of the word $w$ to which it is applied. 

% The delta vector $\delta$ lies in $\mathbb{R}^{1\times768}$ and is repeated to match the length $n_w$ of the word $w$, thus applied to the word embedding $e_w \in \mathbb{R}^{n_w\times768}$.

% For cases where negative content $C$ consists of multiple tokens, we aggregate these tokens into a single vector using mean-pooling, which then serves as the delta vector.

According to Equation~\ref{Applying-Delta}, the delta vector's influence on the word embedding increases when $\alpha > 0$ and decreases when $\alpha < 0$. 
Since our goal is to control the influence of negative content $C$ within the word embedding $e_w$, the delta vector $\delta$ should be oriented toward the direction representing this content.

\subsection{Suppression on Stable Diffusion}
 We propose an efficient zero-shot method for obtaining the delta and demonstrate its effective application within the cross-attention mechanism of Stable Diffusion.

\subsubsection{Zero-Shot Approach for Delta Vector}
We aim to obtain a delta vector in text embedding space that reduces the influence of negative content $C$ on the generated image. For this purpose, we analyze the impact of the delta vector on the image generation process.
In Stable Diffusion, the prompt $P$ is encoded into a text embedding $\hat{e}$ %= \xi(\tau(P))$, 
which is then projected into key and value features of cross-attention through $f_K$ and $f_V$.
When the delta vector is applied to the word embedding $e_w \in \hat{e}$, %then the value feature can be expressed as:
as $f_V$ is a linear function, the values feature can be expressed as follows:
\begin{equation} 
\begin{aligned} 
f_V(e_{w} + \alpha_v \delta) = f_V(e_w) + \alpha_v f_V(\delta),%, \quad (\alpha_v < 0),
\end{aligned} \label{delta_attention_value}
\end{equation}
where $\alpha_v$ is the modification coefficient that is applied to the value feature.
Since the value feature provides conditions for the generated image, the image can be modified by controlling the delta vector.
Specifically, when $\alpha_v$ is set to a negative value, the influence of the feature $f_V(\delta)$ is weakened in the generated image.
By defining delta as the embedding of negative content $C$, it can effectively suppress the features responsible for generating the negative content. For instance, in Figure~\ref{fig2}, when the delta is defined as the embedding of ``glasses", it is projected into the diffusion feature space, acting as a feature that weakens Steve Jobs' glasses (the actual attention map is shown in Suppl.6.3). In this example, the delta is represented as follows: 
\begin{equation}
    \begin{aligned}
        \widetilde{e}=\left[ e_0, \cdots, e_{\text{glasses}} ,\cdots ,e_{M}\right]&=\xi(\tau(\text{glasses}))\\
        \delta:=e_{\text{glasses}}&.
    \end{aligned}
\end{equation}
For cases where negative content $C$ consists of multiple tokens, we aggregate these tokens into a single vector using mean-pooling, which then serves as the delta vector.
This approach enables the acquisition of a delta vector in a zero-shot manner without requiring additional training.

Additionally, when delta is applied to the key feature, it can be interpreted as follows:
\begin{equation} 
\begin{aligned} 
\frac{Q \cdot \big(f_K(e_{w} + \alpha_k \delta)\big)^\top}{\sqrt{d_k}}=&\frac{Q \cdot \big(f_K(e_{w})\big)^\top}{\sqrt{d_k}}\\ &+ \alpha_k \frac{Q \cdot \big(f_K(\delta)\big)^\top}{\sqrt{d_k}},
\end{aligned} \label{delta_attention}
\end{equation}
% \begin{equation} 
% \begin{aligned} 
% \frac{Q \cdot \big(f_K(e_{w} + \alpha_k \delta)\big)^\top}{\sqrt{d_k}}=\frac{Q \cdot \big(f_K(e_{w})\big)^\top}{\sqrt{d_k}}+ \alpha_k \frac{Q \cdot \big(f_K(\delta)\big)^\top}{\sqrt{d_k}},
% \end{aligned} \label{delta_attention}
% \end{equation}
where $\alpha_k$ represents the modification coefficient applied to the key feature.
The dot product between the query and key $f_K(\delta)$ serves as an attention map, indicating the region where the condition will be generated.
Thus, if the delta is the content embedding, applying it can identify the region where the content is generated in the resulting image.

\subsubsection{Selective Suppression with Delta Vector}
We next describe a method to adapt the delta vector to the cross-attention mechanism for selective suppression. 
We concatenate an additional token to the target token $w$, whose embedding is modified by the delta to become $e^*_{w}$. Specifically, depending on whether the embedding is projected as a key or a value, the delta is applied in the opposite direction.
% As discussed in previous studies~\cite{li2023stylediffusion,p2p}, the key \( K \) attends to the regions where the condition is expected to appear in the query \( Q \), and the value \( V \) provides the features needed to apply the condition to the generated image, with the attention map being calculated using softmax which makes tokens compete in a token-wise manner. To leverage this property, we concatenate an additional token, whose embedding is modified by the delta to become $e_w^*$, into the original prompt sentence.
We will denote the text embeddings used to construct $K$ and $V$ as $\hat{e}_k$ and $\hat{e}_v$, respectively, and refer to the modified embeddings as $e_k^*$ and $e_v^*$. This approach leverages the normalization effect of the softmax function: $e_k^*$ competes with the target embedding $e_w$, precisely locating where the negative content would appear. Then $e_v^*$ suppresses the visual feature of the content corresponding to those regions.
% This approach effectively reduces attention toward the embeddings of other words, including the original embedding $ e_w $ that encodes negative content $C$, thus preventing it from being emphasized in the output (e.g., as shown in Figure~\ref{fig2}, the attention map of $ e_w $). 
Our method not only suppresses negative content within the target regions, but also prevents tokens entangled with negative content from inadvertently generating it.

To selectively control the suppression of potential negative content regions for \( C \), we define modified embeddings \( e_k^* \) and \( e_v^* \). First, we formulate the updated key \( K^* \) and embedding \( e_k^* \) as follows:
\begin{equation} 
\begin{aligned} 
e_{k}^*& = e_w + \alpha_k\delta, \quad (\alpha_k > 0),\\
K^* = f_K(\hat{e}_k)&, \quad \text{ where }  \hat{e}_k = \left[ e_0, \dotsb, e_w, e_k^*, \dotsb ,e_{M} \right] , 
\label{apply_key}
\end{aligned} 
\end{equation}
\noindent where delta is added to word embedding \(e_w\), it produces \(e_k^*\), which emphasizes the content $C$.
Thus, the attention map of the embedding \(e_k^*\) will strongly attend to the regions where negative content $C$ is expected to be generated. 
Once such regions are identified to prevent $C$ from appearing in these regions, we provide a feature that suppresses $C$ within the attended regions of \(e_k^*\).
Therefore, we formulate a new value \(V^*\) and the modified embedding ${e}_v^*$ as follows:
\begin{equation} 
\begin{aligned} 
e_v^*& = e_w + \alpha_v\delta, \quad (\alpha_v < 0),\\
V^* = f_V(\hat{e}_v)&, \quad \text{ where }  \hat{e}_v = \left[ e_0, \dotsb, e_w, e_v^*, \dotsb ,e_{M}\right] , 
\label{apply_value}
\end{aligned} 
\end{equation}
where subtracting the delta from \(e_w\) yields \(e_v^*\), which weakens the content $C$.
This approach effectively suppresses the generation of negative content $C$ in the image.

Finally, the cross-attention of our method can be defined as follows:
\begin{equation}
    \label{eq1}
    \begin{aligned}
    \text{Attention}(Q, K^*, V^*) = \text{softmax}\left(\frac{Q \cdot K^{*\top}}{\sqrt{d_k}}\right) \cdot V^*.
    \end{aligned}  
\end{equation}
% In this way, we can selectively suppress negative content $C$ within the targeted regions while ensuring that other areas remain unaffected.
By operating directly at the embedding level, the proposed competitive token mechanism enables precise and localized suppression, effectively addressing the challenge of entangled content.

\subsection{Suppression on Personalized T2I Model}
Suppression becomes challenging in models fine-tuned with personalized techniques~\cite{dreambooth, customdiffusion} due to weakened prior knowledge and strong overfitting to specific subjects.
For instance, when a personalized T2I model is fine-tuned to generate ``$S^*$ dog" (a specific dog wearing glasses, depicted on the right side of Figure~\ref{fig2}), the model tends to learn to generate the glasses as part of the subject's identity.
Since personalized T2I models generally aim to preserve the identity of ``$S^*$", they overfit the subject appearance, making it challenging to suppress ``glasses."
Moreover, as mentioned in Section 3.2, the personalized T2I model struggles to retain prior information during the fine-tuning process. As a result, the prior information about ``glasses" becomes weakened, making it even more difficult to identify and suppress them effectively.
%Moreover, due to the weakened prior information about ``glasses," it becomes hard to identify and suppress them effectively.
Thus, our SSDV method with a zero-shot delta inaccurately suppresses ``glasses", when the prompt ``a photo of $S^*$ dog" is given.
To address this issue, we propose an optimization-based approach to obtain a delta that accurately captures the negative content entangled with $S^*$, enabling effective suppression in personalized T2I models.

\subsubsection{Optimization Approach for Delta Vector}
Our optimization approach to obtain the delta requires the subject image $I$ and the mask $M$, indicating the negative content region in $I$.
To align the delta 
$\delta$  with the negative content, we define the modified embedding as follows:
% To make delta $\delta$ correspond to the negative content, we provide a modified embedding as the following equation:
\begin{equation} 
\begin{aligned}
e_k^* = e_w + \alpha_k\delta, \quad (\alpha_k > 0), \\
e_v^* = e_w + \alpha_v\delta, \quad (\alpha_v > 0),
\label{optim_delta_modify}
\end{aligned} 
\end{equation}
%where this modified embedding $e_k^*$ and $e_v^*$ are each used for key $K$ and value $V$ during training.
where $e_{k}^*$ and $e_{v}^*$ are computed following the procedure described in Equation (\ref{apply_key}) and (\ref{apply_value}), respectively.
During training, $\alpha_k$ and $\alpha_v$ are set to a positive value to optimize the delta to capture negative content well within the image.
The training process is guided by the following objective function:
\begin{equation} 
\begin{aligned} 
\mathcal{L}_{latent}(\delta) =  \frac{\sum_{t=1}^{T} \sum_{h,w} M_{h,w} \, |z_{t-1}^{inv} - \hat{z}_{t-1}|_{h,w}}{\sum_{h,w} M_{h,w}},
\label{latent_loss}
\end{aligned}
\end{equation}
\(z_{t-1}^{\text{inv}}\) is the inverted latent on timestep ${t-1}$ obtained by inverting image $I$ with DDIM inversion~\cite{song2020denoising}.
$\hat{z}_{t-1}$ is the predicted latent of each timestep ${t-1}$ obtained during the reverse process, as shown in the following equation:
\begin{equation} 
\begin{aligned} 
\hat{z}_{t-1} = \frac{1}{\sqrt{\alpha_t}} \left( z_t - \frac{1 - \alpha_t}{\sqrt{1 - \bar{\alpha}_t}} \epsilon_\theta (z_t, t) \right),
\label{reverse_process}
\end{aligned} 
\end{equation}

\noindent where \(\alpha_{t}\) and $\bar{\alpha}_t$ defined as $\prod_{\tau=1}^t \alpha_\tau$ are the parameters for DDIM scheduling.
By training delta using the objective function $\mathcal{L}_{latent}$, the negative content generated by $e_k^*$ and $e_v^*$ aligns with the negative content in the image $I$, allowing delta to accurately capture the negative content.

Additionally, to ensure that the $e_k^*$ effectively attends to the region of negative content, we train a delta vector using the following objective function:
\begin{equation}
\begin{aligned}
\mathcal{L}_{attn}(\delta) &= \sum_{t=1}^{T}|M - \hat{M}_{t-1}|\label{atten_loss},
\end{aligned} 
\end{equation}
where \(\hat{M}_{t-1}\) is the attention map from cross-attention corresponding with \(e_k^*\). 
This attention map \(\hat{M}_{t-1}\) is derived from the first up-sampling block, which is known to effectively capture the semantic mask~\cite{emerdiff, openvocadiff, luo}.
This $\mathcal{L}_{attn}$ provides that $e_k^*$ correctly attends to the negative content region in cross-attention.
Therefore, the overall objective function can be defined as follows:
\begin{equation} 
\begin{aligned} 
\mathcal{L}_{optim} = \mathcal{L}_{latent} + \lambda_{attn} \mathcal{L}_{attn} ,
\end{aligned} \label{total_loss}
\end{equation}
where \(\lambda_{attn}\) is the weighting coefficient for $\mathcal{L}_{attn}$.

\subsection{Local Blending for Preserving Original}
% 한편 원본 이미지의 구조를 유지하기 위해 우리는 attention feature blending를 제안한다. attention feature blending은 \text{Attention}(Q, K^*, V^*)와 \text{Attention}(Q, K, V)의 차이를 구하고 일정 수준 이상의 차이만을 반영하는 blending 방법이다
Meanwhile, controlling the image by modifying only the text embeddings often struggles to preserve the structure of the original image. To address this issue, we propose blending at the attention feature level. Specifically, we generate a threshold-based binary mask from the difference between \(\text{Attention}(Q, K^*, V^*)\) and \(\text{Attention}(Q, K, V)\), selectively combining only significant differences. 

Furthermore, inspired by the methods proposed in~\cite{p2p, sega}, we design latent-level blending by combining the attention-map-based binary mask~\cite{p2p} and the binary mask derived from differences~\cite{sega} before and after suppression, thus achieving more precise suppression. For further details, refer to Suppl.3.
\section{Experiments}

\begin{table*}[!ht]
   \centering
   \scalebox{0.75}{
    \begin{tabular}{lccccccccccccc}
        \toprule
        & \multicolumn{4}{c}{Suppression in SD} & \multicolumn{9}{c}{Suppression in DreamBooth-based model } \\
        \cmidrule(lr){2-5} \cmidrule(lr){6-14}
        Method & \multicolumn{4}{c}{SEP-Benchmark} & \multicolumn{3}{c}{\textit{S*} bowl $-$ berry} & \multicolumn{3}{c}{\textit{S*} figure $-$ cape} & \multicolumn{3}{c}{\textit{S*} toy duck $-$ yellow} \\
        \cmidrule(lr){2-5} \cmidrule(lr){6-8} \cmidrule(lr){9-11} \cmidrule(lr){12-14}
        & CLIP↓ & IFID↑ & DetScore↓ & Prefer↑ & CLIP↓ & IFID↑ & Prefer↑ & CLIP↓ & IFID↑ & Prefer↑ & CLIP↓ & IFID↑ & Prefer↑ \\
        \midrule
        SD (Generated image) & 17.60 & 0.0 & 0.314 & - & 24.32 & 0.0 & - & 17.04 & 0.0 & - & 17.20 & 0.0 & - \\
        \midrule
        Negative prompt & \underline{16.46} & 40.34 & 0.265 & \underline{4.54} & 20.15 & 41.81 & 2.27 & 17.00 & 37.73 & 0.75 & 16.58 & 17.67 & \underline{3.03}\\
        P2P~\cite{p2p} & 18.34 & 9.65 & 0.392 & 2.15 & \underline{16.09} & 183.51 & 1.51 & 17.07 & 25.38 & \underline{3.03} & 16.96 & 5.12 & 0.75\\
        SuppressEOT~\cite{suppresseot} & 18.41 & 45.89 & 0.389 & 1.36 & 18.70 & \underline{212.05} & 2.27 & 18.14 & \underline{89.27} & 1.15 & 16.84 & 44.26 & 0.0 \\
        SEGA~\cite{sega}  & 17.16 & 15.34 & 0.298 & 3.52 & 21.53 & 35.93 & 1.51 & \underline{16.30} & 14.03 & 0.75 & 17.04 & 12.89 & 0.0 \\
        Inst-Inpaint~\cite{Instinpaint} & 17.14 & \underline{47.50} & \underline{0.220} & 1.93 & 22.54 & 2.27 & \underline{8.63} & 17.03 & 29.23 & \underline{3.03} & \textbf{12.22} & \textbf{279.61} & 0.75 \\
        % Ours & \textbf{16.17} & \textbf{72.63} & \textbf{0.012} & \textbf{86.47} & \textbf{14.14} & \textbf{221.50} & \textbf{90.15} & \textbf{15.65} & \textbf{118.24} & \textbf{90.90} & \underline{14.25} & \underline{118.26} & \textbf{95.45} \\
        Ours & \textbf{15.92} & \textbf{87.34} & \textbf{0.113} & \textbf{86.47} & \textbf{14.14} & \textbf{221.50} & \textbf{90.15} & \textbf{15.65} & \textbf{118.24} & \textbf{90.90} & \underline{14.25} & \underline{118.26} & \textbf{95.45} \\
        \bottomrule
    
    \end{tabular}
    }
    \vspace{-5pt}
    \caption{Comparison with other previous methods. The best results are in \textbf{bold}, and the second best results are \underline{underlined}.}
    \label{tab:quantitative_table}
\end{table*}

\subsection{Experimental Details}
\paragraph{Datasets} To evaluate our method for entangled content suppression assessment, we developed the Strongly Entangled Prompts Benchmark (SEP-Benchmark). 
To construct the benchmark, we collected cases where specific prompts consistently generate entangled content in Stable Diffusion, regardless of seed.
For instance, the prompt ``A photo of a bed" almost always generates an image with a ``pillow". 
Similar to recent related works~\cite{null, erasing, localizing, suppresseot}, which used approximately 1000 images for evaluation, we collected 10 prompt-content pairs and generated 100 images for each prompt using different random seeds. We validate the strong entanglement observed in the prompt pairs of the SEP-benchmark by measuring their entanglement. The details of this are provided in Suppl.5.
To further evaluate our proposed method on models fine-tuned with personalized techniques~\cite{dreambooth, customdiffusion}, we selected unique subjects with strongly entangled content from DreamBooth~\cite{dreambooth} and VICO~\cite{vico}.
For each subject, we generated 1000 images using different random seeds.
% to assess content suppression.
% \begin{figure}[!ht]
%     \centering
%     \includegraphics[width=0.4\textwidth]{figure/human_preference.pdf} 
%     \caption{Human preference comparison at DreamBooth tuned model. The left section is the score of our method and the right section is the score of other methods.}
%     \label{fig:human_preference}  
% \end{figure}
\vspace{-12pt}
\paragraph{Metrics} To evaluate content suppression effectiveness, we use \textbf{CLIP}~\cite{clipscore} where lower cosine similarity between CLIP features of the image and the negative prompt indicates successful suppression. 
Additionally, \textbf{Fréchet Inception Distance (FID)}~\cite{fid} was used to measure the differences between image distributions before and after suppression, with a higher FID indicating stronger suppression (referred to as IFID)~\cite{suppresseot}.
We also employed \textbf{DetScore} which is based on MMDetection~\cite{MMDetection}, an open-vocabulary object detection network. 
Suppression was considered successful when dectector failed to detect the negative target.
Furthermore, we conducted a user study with 50 participants, evaluating their \textbf{Preference} (referred to as Prefer) for suppressed images. The details of our user study are in Suppl.9. 
% Furthermore, a \textbf{user study} was conducted to demonstrate that our method outperforms baselines, involving 50 participants who were asked to select the better suppression result between our method and the baselines.

% \vspace{-6pt} 
% \paragraph{Implementation details} In our implementation, we utilized the Stable Diffusion v1.5\cite{high}. In all our experiments, we set $\alpha_k = 1.3$ and $\alpha_v = -1.3$, respectively. For delta optimization experiment, $\lambda_1$, $\lambda_2$, and $\alpha$ were all set to 1.0.

\subsection{Experimental Results}
\vspace{-5pt}
We compare our methods with the following baselines: Negative Prompt, Prompt-to-Prompt~\cite{p2p}, SuppressEOT~\cite{suppresseot}, SEGA~\cite{sega}, and Inst-Inpaint~\cite{Instinpaint}. We conducted experiments using the default settings provided in the official code of each baseline. 
In baselines such as P2P and SuppressEOT, the negative content word must be explicitly included in the prompt, typically formatted as “... without (negative content)”.
To evaluate the suppression performance on SD, we examined the results of our SSDV using the zero-shot approach.
Additionally, we conducted experiments on personalized models to demonstrate the performance of both zero-shot and optimization-based delta approaches.
We used DreamBooth~\cite{dreambooth} as the personalized technique, with additional results using CustomDiffusion~\cite{customdiffusion} provided in Suppl.7.
\begin{figure*}[!ht]
    \centering
    \vspace{-11pt} 
    \includegraphics[width=0.97\textwidth]{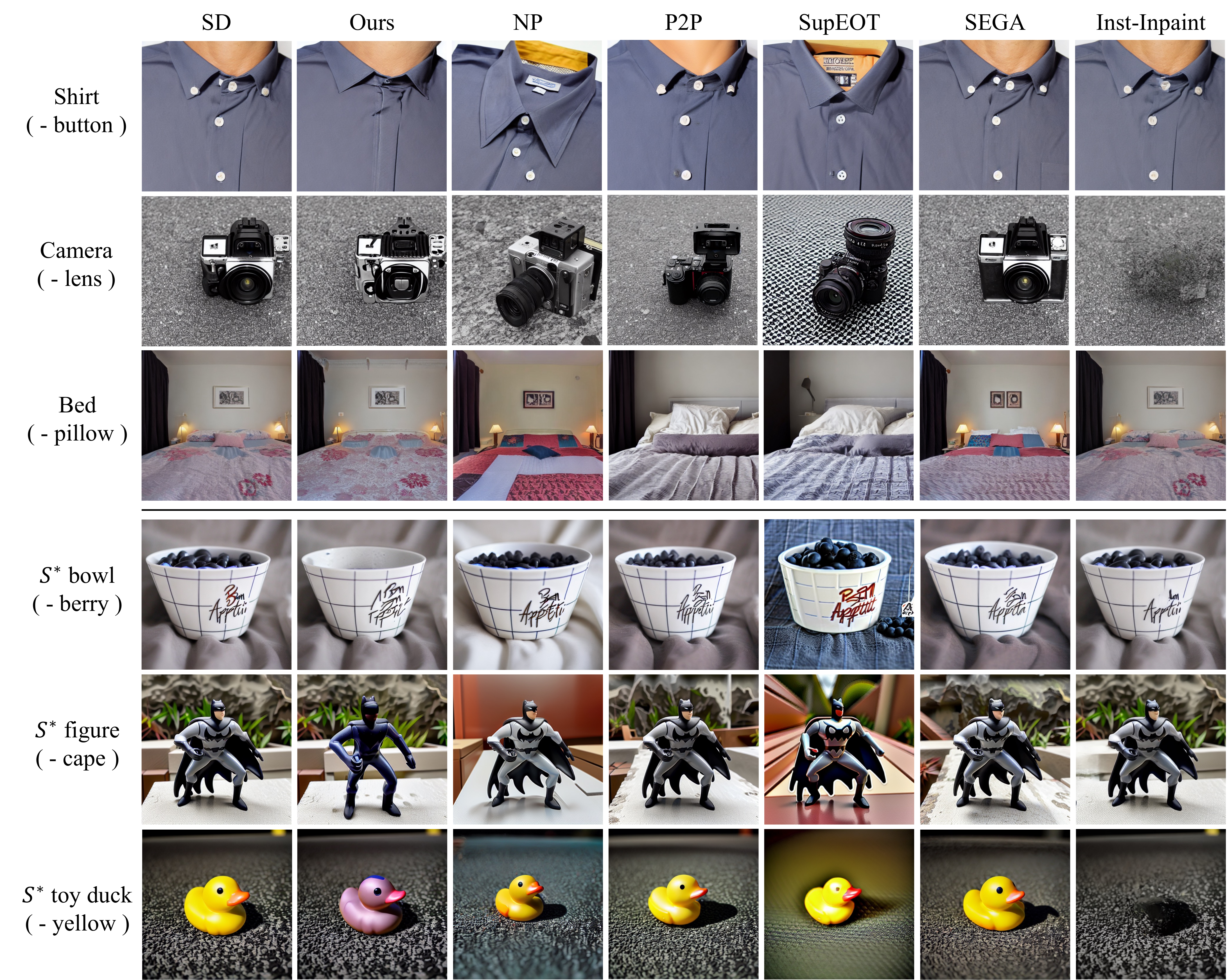} 
    \vspace{-7pt}
    \caption{Qualitative results with other methods: (Top) Stable Diffusion, (Bottom) DreamBooth-based. We can suppress the content (which is in the leftmost parentheses) when the input prompt (located above parentheses) is given.}
    \label{result}  
    \vspace{-9pt}
\end{figure*}

\subsubsection{Suppression in SD}
% \paragraph{Suppression in SD}
As reported in Table~\ref{tab:quantitative_table}, our method outperformed all baselines for suppression. 
Although NP achieve the second-best scores, the suppression by NP is incomplete, as shown in Figure~\ref{result} (third column, top). 
In the case of Inst-Inpaint which achieves the third-best scores, as observed in Figure~\ref{result} (second row, seventh column), not only is the lens suppressed, but the related item, the camera, is also removed. 
Other methods, such as SEGA, SuppressEOT and P2P failed to suppress the content and, in some cases, also altered the identity of the original image.
In contrast, our method achieves complete suppression of strongly entangled content, as shown in Figure~\ref{result} (second column, top). 
Furthermore, as demonstrated in Table~\ref{tab:quantitative_table} preference result, the user study indicates a strong preference for our method over other baselines in terms of suppression performance.
% \sout{Furthermore, in the user study, 86\% of participants preferred our method, which is 81.93\% higher than the second-best method, NP.}
% 심지어 user study에서 86%의 사람들이 our methods를 선호하였다 (두번째로 좋은 NP보다 81.93%나 더 높다.)
% However, we note that in Figure~\ref{result} (fourth row, second column), the identity of Steve Jobs is slightly compromised. This can be addressed by adjusting the alpha value to better balance suppression with identity preservation.
% This issue arises from using a high alpha $\alpha$ value. By adjusting the alpha, it is possible to achieve effective suppression while preserving the identity as much as possible.

% \begin{table}[H]
%    \centering
%    \resizebox{\linewidth}{!}{ % 가독성 유지하면서 표 크기 자동 조정
%     \begin{tabular}{lcccc}
%         \toprule
%         \multirow{2}{*}{Method} & SEP- & \multicolumn{3}{c}{DreamBooth-based Suppression} \\
%         & Benchmark & S* bowl $-$ berry & S* figure $-$ cape & S* toy duck $-$ yellow \\
%         \midrule
%         Negative prompt & \underline{4.54} & 2.27 & 0.75 & \underline{3.03} \\
%         P2P~\cite{p2p} & 2.15 & 1.51 & \underline{3.03} & 0.75 \\
%         SuppressEOT~\cite{suppresseot} & 1.36 & 2.27 & 1.15 & 0.0 \\
%         SEGA~\cite{sega}  & 3.52 & 1.51 & 0.75 & 0.0 \\
%         Inst-Inpaint~\cite{Instinpaint} & 1.93 & \underline{8.63} & \underline{3.03} & 0.75 \\
%         Ours & \textbf{86.47} & \textbf{90.15} & \textbf{90.90} & \textbf{95.45} \\
%         \bottomrule
%     \end{tabular}
%     }
%     \caption{Comparison of human preference scores across baselines.}
%     \label{tab:prefer_table}
% \end{table}

\subsubsection{Suppression in Personalized T2I Model}
\begin{figure}[ht]
    \centering
    \includegraphics[width=0.45\textwidth]{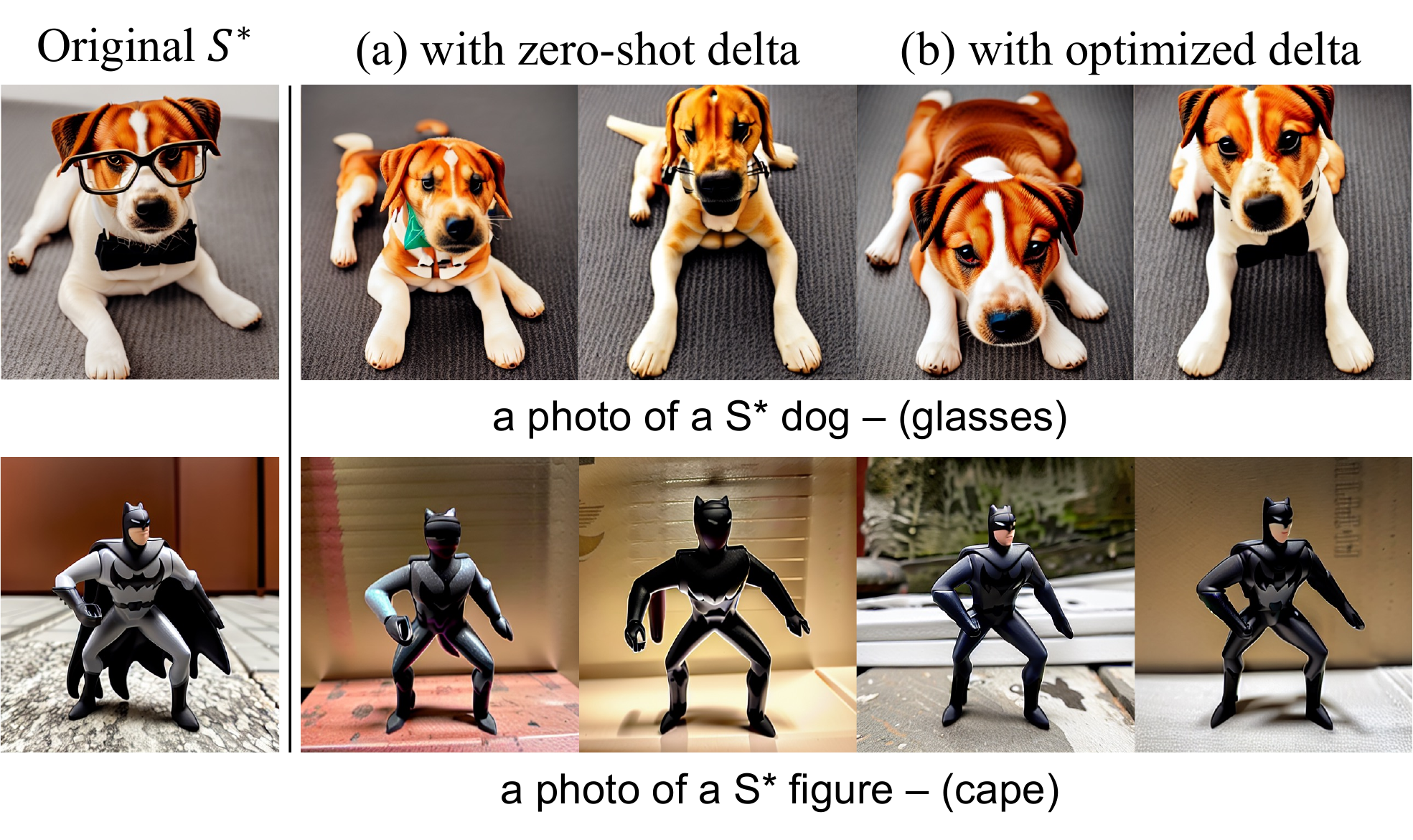} 
    \vspace{-10pt} 
    \caption{Suppression results on the model fine-tuned for specific subject \(S^*\) using DreamBooth: (a) using zero-shot delta, and (b) using optimized delta. }
    \label{delta_optim}  
\end{figure}

\paragraph{Zero-shot approach}
In Table~\ref{tab:quantitative_table}, the `$S^*$ bowl - berry' entry represents the suppression of ``berry" from the prompt ``a photo of $S^*$ bowl" (same for `$S^*$ figure - cape' and `$S^*$ toy duck - yellow'). 
The originally generated result for ``$S^*$ bowl" is shown in the first column of the fourth row in Figure~\ref{result}, where it is evident that the content ``berry" is strongly entangled with ``$S^*$ bowl".
As shown in the second column of the fourth row in Figure~\ref{result}, our method successfully suppresses the berries, while the remaining columns indicate that the baselines fail to do so. 
Additionally, as presented in Table~\ref{tab:quantitative_table}, our method outperforms all baselines in terms of CLIP and IFID scores for both `$S^*$ bowl - berry' and `$S^*$ figure - cape', except for `$S^*$ toy duck - yellow', where Inst-Inpaint achieves the best performance.
However, as demonstrated in the last column of the last row in Figure~\ref{result}, this occurs because Inst-Inpaint suppresses not only the negative content ``yellow" but also the subject ``$S^*$ toy duck", resulting in better CLIP and IFID scores.
In contrast, our method (second column of the last row in Figure~\ref{result}) effectively suppresses only the yellow color while preserving the toy duck, demonstrating better suppression performance. This is further supported by the human preference scores, where Inst-Inpaint received 0.75\%, whereas our method achieved 95.45\%.
% , indicating a significantly better performance.
% \sout{This is further supported by the human preference scores, where Inst-Inpaint received 0.75\%, whereas our method achieved 95.45\%, indicating a significantly better performance.}
% A similar pattern is observed in the fifth row of Figure~\ref{result} for the `figure - cape' case.
\vspace{-12pt}
\paragraph{Optimization approach}
% delta optimizing approach가 zero-shot approach에 비해 정확한 delta를 잘 capture하는지 확인하기 위해, 정성적 비교와 user study를 진행하였다.\
We conducted qualitative comparisons and a user study to evaluate whether the delta optimization approach more accurately captures content than the zero-shot approach.
In the first row of Figure~\ref{delta_optim}, applying the optimized delta vector (Figure~\ref{delta_optim}.(b)), effectively suppresses ``glasses" while maintaining the identity of the original subject $S^*$. 
In contrast, Figure~\ref{delta_optim}.(a) uses a zero-shot delta, which also suppresses ``glasses", but is less effective in preserving identity.
These results indicate that optimizing the delta vector enables the model to more accurately capture the content to be suppressed in DreamBooth-tuned models.
Additionally, the user study in Table~\ref{tab:delta_optim} shows that 73.2\% of participants preferred the optimization-based delta approach for better identity preservation and more effective content suppression. 
\vspace{-10pt}
\vspace{-17pt}
\begin{table}[H]
    \centering
    \scalebox{0.75}{
    \begin{tabular}{l|ccc}
    \toprule
    Method & Zero-shot Delta & Optimized Delta & Difference (\%) \\
    \midrule
    Preference (\%) ↑ & 26.8 & \textbf{73.2} & +46.4 \\
    \bottomrule
    \end{tabular}
    }
    \vspace{-5pt}
    \caption{Human preference in the comparison between zero-shot delta and optimized delta.}
    \label{tab:delta_optim}
\end{table}
\vspace{-15pt}
\subsection{Ablation Study}
\begin{figure}[t]
    \centering
    \includegraphics[width=0.33\textwidth]{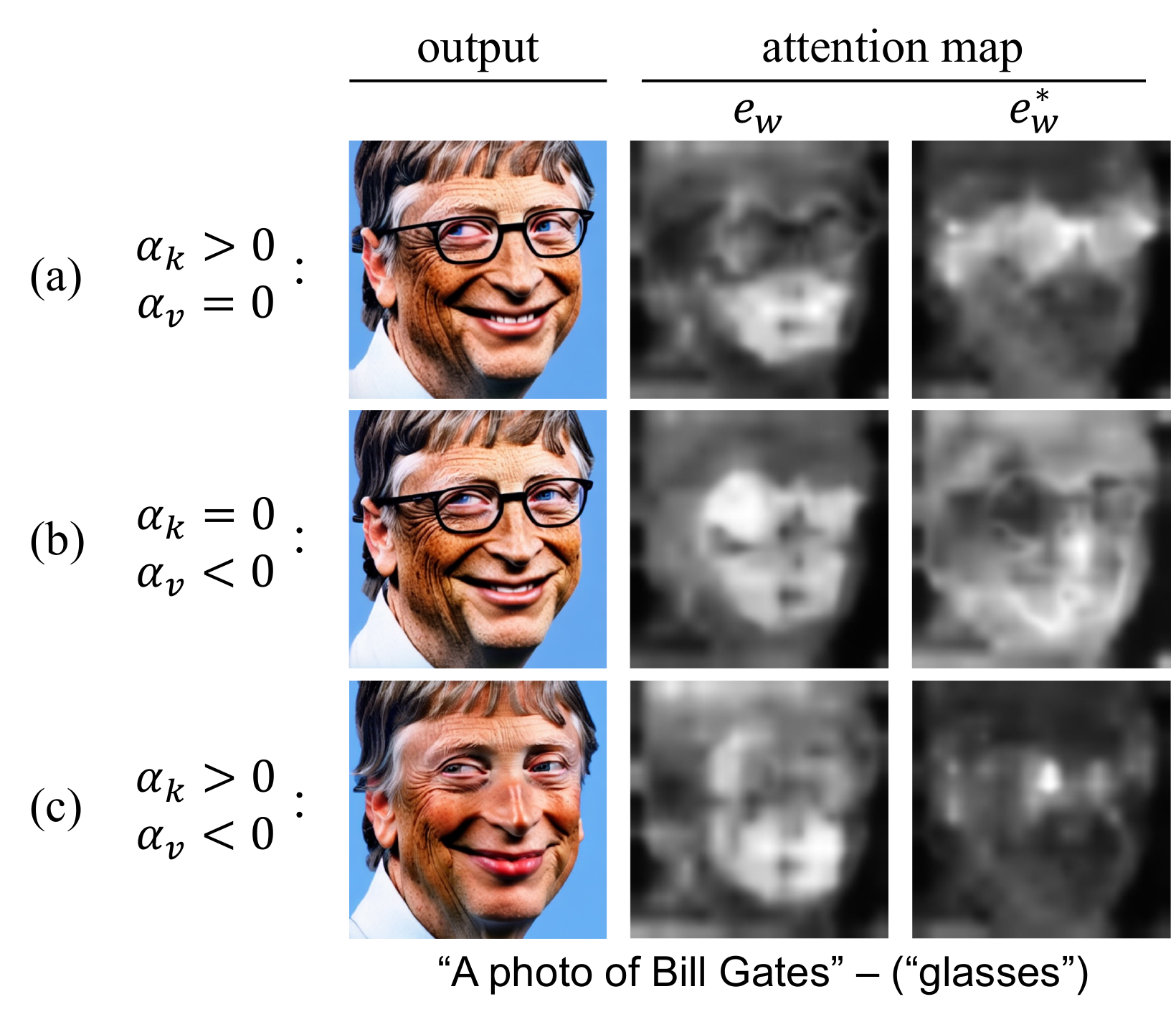} 
    \vspace{-10pt} 
    \caption{Ablation results for our SSDV. %applying the delta to the key and value features.
    The leftmost values represent $\alpha_k$ and $\alpha_v$, while the figures on the right show the output images and the attention maps of the embeddings for each case.
    }
    \label{fig:ablation}  
\end{figure}

We conducted an ablation study to understand the impact of applying the delta vector to key and value features in the SSDV method. The results of the ablation study are shown in Table~\ref{tab:ablation} and Figure~\ref{fig:ablation}.
Figure~\ref{fig:ablation}.(a) shows the result when the delta is applied only to the key feature in a positive direction. 
In this case, the region of the negative content is strongly attended in the attention map of $e_w^*$, which prevents the region from being attended by $e_w$ that is strongly entangled with the content.
% This is due to the competitive nature of the softmax operation in cross-attention.
In contrast, when the delta was not applied to the key feature, as shown in Figure~\ref{fig:ablation}.(b), the region of negative content was attended in the attention map of $e_w$. 
This demonstrates that applying the delta to the key feature effectively prevents other tokens entangled with content from attending to content regions.
However, as shown in the output image of Figure~\ref{fig:ablation}.(a), the content is still generated because $e_w^*$ contains the content information in the value feature. 
\begin{table}[t]
    \vspace{-5pt}
    \centering
    \scalebox{0.8}{
    \begin{tabular}{l|ccc}
    \toprule
    \textbf{$\alpha_k$} ~ \textbf{$\alpha_v$} & ~~CLIP↓~~ & ~~IFID↑~~ & ~~OwlViT↓~~ \\
    \midrule
    ~+ ~~ 0.0 & 17.53 & 27.49 & 0.049 \\
    0.0 ~~ - & 17.08 & 23.01 & 0.037 \\
    ~+ ~~~~ - & \textbf{16.17} & \textbf{72.63} & \textbf{0.012} \\
    \bottomrule
    \end{tabular}
    }
    \vspace{-4pt}
    \caption{Quantitative results from the ablation study comparing each case of applying the delta to key and value features.}
    \label{tab:ablation}
    \vspace{-16pt}
\end{table}
When the delta was applied in the negative direction to the value, Figure~\ref{fig:ablation}.(c) shows that the content is successfully suppressed.
This demonstrates that applying the delta in the negative direction to the value feature effectively suppresses the content in the image.
In Figure~\ref{fig:ablation}.(b), applying the delta only to the value feature results in ineffective suppression, indicating that applying the delta to both the key and value (in positive and negative directions, respectively) is necessary for successful content suppression.
% Figure~\ref{fig:ablation}.(c) demonstrates effective suppression by directing attention to the region where the content would be generated and providing the required suppression information.
Although the attention map of $e_w^*$ in Figure~\ref{fig:ablation}.(c) may appear unclear, this is simply because it was visualized during the suppression process.
In conclusion, applying the delta to both key and value features is crucial for effective content suppression, as also evidenced by the best performance results presented in Table~\ref{tab:ablation}.

% (where $\alpha_k$ is positive, and $\alpha_v$ is negative).

% By comparing cases (a) and (d), the significance of the value in the attention mechanism becomes evident. Specifically, in the attention map of (d), adding delta to $e_w^*$ leads to the suppression of the glasses in $e_w$. This effect is attributed to the softmax operation in the attention map, which causes tokens to compete for activation. Leveraging this property, we can indirectly suppress the content in other tokens. This is further demonstrated by comparing (a) and (c), where (a) effectively suppresses content by adding delta indirectly, whereas (c), which directly subtracts delta, cannot.

% In summary, the comparison between the results of (a) and (d) confirms that applying delta to the value is crucial, and the comparison among (a), (b), and (c) shows that adding delta to the key is a more effective strategy for content suppression.
\subsection{Additional Results}
In our supplementary material, we present additional qualitative results from SEP-Benchmark. Moreover, we demonstrate the application of our method across various tasks and provide in-depth analyses of our SSDV method, offering insights into its behavior and performance.

\section{Conclusion}
We propose a zero-shot method to suppress the content strongly entangled with the prompt by obtaining the delta vector that modifies text embeddings. Additionally, we introduce SSDV for selective region suppression and extend our method to personalized T2I models through optimization. Our extensive evaluations demonstrate that our approach outperforms existing methods both quantitatively and qualitatively.

% We propose a method to suppress content that is strongly entangled with the prompt. 
% Specifically, we introduce a zero-shot approach to obtain the delta vector that modifies the text embedding for suppression.
% Additionally, we propose the SSDV method that selectively suppresses the content. % while minimizing the impact on other areas.
% We further extend our approach to personalized T2I models with an optimization-based method for more precise suppression.
% Our extensive evaluation results show that our approach outperforms existing methods both quantitatively and qualitatively.

\section*{Acknowledgements}
\vspace{-0.2cm}
This work was supported in part by MSIT/IITP (No. RS-2022-II220688, RS-2025-02263841, RS-2019-II190421, RS-2022-II220688, RS-2025-02263841, RS-2021-II212068), MSIT/NRF (No. RS-2024-00357729), and KNPA/KIPoT (No. RS-2025-25393280).

{
    \small
    \bibliographystyle{ieeenat_fullname}
    \bibliography{main}
}

\end{document}